\documentclass[10pt,twocolumn,letterpaper]{article}

\usepackage{cvpr}
\usepackage{times}
\usepackage{epsfig}
\usepackage{graphicx}
\usepackage{amsmath}
\usepackage{amssymb}
\usepackage{multirow}
\usepackage{caption}
\usepackage{color}
\usepackage{booktabs}
\usepackage{nth}
\usepackage[font=small,labelfont=bf]{caption}

\definecolor{red1}{RGB}{236,0,140}
\definecolor{green1}{RGB}{0,166,81}
\definecolor{purple}{RGB}{102,45,145}

\usepackage[pagebackref=true,breaklinks=true,letterpaper=true,colorlinks,bookmarks=false]{hyperref}

 \cvprfinalcopy 

\def\cvprPaperID{361} 

\ifcvprfinal\fi
\begin{document}

\title{Improving Fully Convolution Network for Semantic Segmentation }

\author{Bing Shuai   \hspace{2em} Ting Liu \hspace{2em} Gang Wang\\
School of EEE, Nanyang Technological University, Singapore\\
{\tt\footnotesize \{bshuai001, liut0016, wanggang\}@ntu.edu.sg}
}


\maketitle

\begin{abstract}
  Fully Convolution Networks (FCN) have achieved great success in dense prediction tasks including semantic segmentation. In this paper, we start from discussing FCN by understanding its architecture limitations in building a strong segmentation network. Next, we present our Improved Fully Convolution Network (IFCN). In contrast to FCN, IFCN introduces a context network that progressively expands the receptive fields of feature maps. In addition, dense skip connections are added so that the context network can be effectively optimized. More importantly, these dense skip connections enable IFCN to fuse rich-scale context to make reliable predictions. Empirically, those architecture modifications are proven to be significant to enhance the segmentation performance. Without engaging any contextual post-processing, IFCN significantly advances the state-of-the-arts on ADE20K (ImageNet scene parsing), Pascal Context, Pascal VOC 2012 and SUN-RGBD segmentation datasets.

\end{abstract}

\section{Introduction}
Fully Convolution Network (FCN) has evolved to be the \emph{de facto} network architecture in semantic segmentation after it is successfully adopted in \cite{long2015fully}.
As it is difficult (usually fails) to train a strong FCN from scratch, Long \emph{et al.} simply adapts the architecture of pre-trained classification network (e.g. VGG-16 \cite{simonyan2014very} trained on ImageNet \cite{russakovsky2015imagenet} etc.).
Even though FCN significantly advances the performance of semantic segmentation, here we discuss its possible drawbacks so that we can build stronger segmentation network architectures.

First and foremost, the pre-trained CNN is trained with low-resolution images (e.g. 224 $\times$ 224 pixels), whereas input segmentation images are usually in high resolution (e.g. $512 \times 512$ pixels).
The simple adaptation techniques adopted in FCN cannot effectively address this domain gap, which leads to less optimized segmentation performance of FCN.
To elaborate, the feature maps that are used for classification in FCN have limited contextual fields,  so predictions are inconsistent for local ambiguous regions.
To close such a domain gap, one option is to adapt (downsample) the inputs. Unfortunately, this practice is found to deteriorate the performance of small-size objects. A more plausible and popular option is to modify network architectures (of FCN) so that they are fit for high-resolution inputs, and we will review and discuss these works in Section \ref{Section:Review}.

Additionally, Long \emph{et al.} \cite{long2015fully} adopts two extra skip connections in FCN-8s to aggregate three scale contextual predictions.
Nonetheless, such limited scale fusion in segmentation networks is not expected to handle the significant variance of object scales across different images.

Next, we present our Improved FCN (IFCN) to address those issues. We take the adaptation of VGG-16 \cite{simonyan2014very} as an example to demonstrate the architecture of IFCN.
\subsection{IFCN vs FCN}
The key modification to FCN is that IFCN plugs in a context network between the pre-trained CNN and upsample layers.
Thus, IFCN entails new parameters, which is crucial to fill the domain gap during the network fine-tuning on high-resolution segmentation images.
The context network is a densely branched convolution network. Specifically, it is stacked with multiple convolution blocks, and shortcut branches are further added from each intermediate feature map.
Functionally speaking, context network is able to significantly expand the receptive fields of feature maps, which is essential to contextualize local semantic predictions.
In addition, shortcut branches are also important to ease the optimization difficulty of context network, as they provide shortcut paths for the propagation of error signals. Meanwhile, those shortcut branches enable IFCN to make predictions based on rich scale contexts. In consequence, IFCN is able to converge to a significantly better local optima than FCN on standard semantic segmentation images.

Besides, IFCN discards the last two \texttt{fc} layers (\texttt{fc6}, \texttt{fc7}) which are specific to image classification \cite{yosinski2014transferable}. By doing this, feature maps are more compact and the network size is reduced as well. We will elaborate the architecture design of context networks in Section \ref{Section:ContextNetwork}.

\subsection{IFCN-xs vs FCN-xs}
We further add more skip connections from early feature maps of pre-trained CNN so that richer scale contexts are incorporated in semantic predictions.
Similar to the notation of FCN-xs, we symbolize our segmentation network as IFCN-xs, where x denotes the sliding stride of the finest-resolution feature maps involved in prediction.
Let's take FCN-8s and IFCN-8s for illustrative comparison.
FCN-8s has two skip connections from feature maps \texttt{pool3} and \texttt{pool4}. IFCN-8s, on the other hand, adds skip connections for all the following feature maps starting with \texttt{pool3}.  In this perspective, IFCN-xs leverages significantly richer scale context for predictions than FCN-xs.  As each feature map carries some distinct scale information, they are complementary to collectively handle the scale variance of objects. It is important to note that these skip connections incorporate very limited computation overhead.  We will present the details of IFCN-xs in Section \ref{Section:IFCN-8s}.

With these architecture modifications, IFCN marries context aggregation as well as rich-scale contextual prediction in an elegant framework.
IFCN delivers a strong segmentation network architecture and it achieves new state-of-the-arts on standard semantic segmentation datasets including
ADE20K (ImageNet Scene Parsing) \cite{zhou2016semantic}, Pascal Context \cite{mottaghi_cvpr14}, VOC 2012 \cite{everingham2010pascal}  and SUN-RGBD \cite{song2015sun}.
In Section \ref{Section:ArchitectureEvaluation}, we also present detailed study to the architecture design of IFCN. Overall, IFCN doesn't involve any new computation layers comparing to FCN. Thus, it can be easily implemented with current deep learning libraries. We will release the code and model upon the acceptance of this paper.

\begin{figure*}
\begin{center}
  \includegraphics[width = 0.975\textwidth]{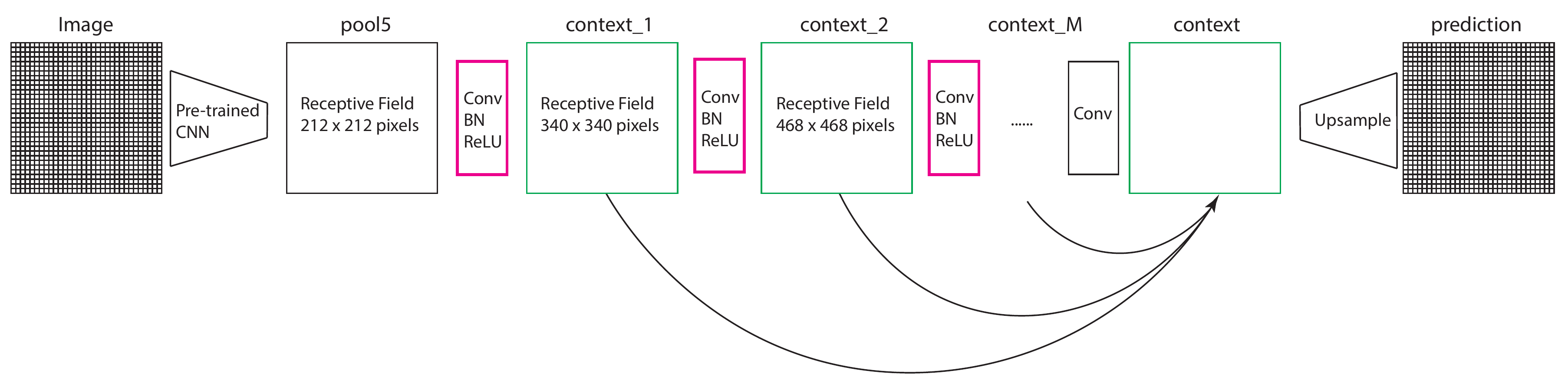}
\end{center}
\caption{Architecture of the \textbf{context network}, which is a densely branched convolution network (a stack of $M$ \texttt{conv} blocks). After each \texttt{conv} block ($5 \times 5$ \texttt{Conv} kernel in the above example), the receptive fields of feature maps are expanded. Note that the spatial dimensionality of these feature maps keeps unchanged.  }
\label{Figure:ContextNetwork}
\end{figure*}

\begin{table*}
\footnotesize
\centering
\begin{tabular}{|r|c|c|c|c|c|c|c|c|c|}
\hline
Layer & \texttt{conv1\_1} & \texttt{conv1\_2} & \texttt{pool1} & \texttt{conv2\_1} & \texttt{conv2\_2} & \texttt{pool2} & \texttt{conv3\_1} & \texttt{conv3\_2} & \texttt{conv3\_3} \\
\hline
Receptive field (px) & $3 \times 3$ & $5 \times 5$ & $6 \times 6$ & $10 \times 10$ & $14 \times 14$ & $16 \times 16$ & $24 \times 24$ & $32 \times 32$ & $40 \times 40$ \\
\hline
\end{tabular}
\begin{tabular}{|c|c|c|c|c|c|c|c|c|c|}
\hline
 \texttt{pool3} & \texttt{conv4\_1} & \texttt{conv4\_2} & \texttt{conv4\_3} & \texttt{pool4} & \texttt{conv5\_1} & \texttt{conv5\_2} & \texttt{conv5\_3} & \texttt{pool5} & \texttt{fc6} \& \texttt{fc7} \\
\hline
 $44 \times 44$ & $60 \times 60$ & $76 \times 76$ & $92 \times 92$ & $100 \times 100$ & $132 \times 132$ & $164 \times 164$ & $196 \times 196$ & $212 \times 212$ & $404 \times 404$   \\
\hline
\end{tabular}
\caption{Receptive fields of feature maps in VGG-16.}
\label{Table:ReceptiveField-VGG-16}
\end{table*}

\section{Review of FCN and its developments}
\label{Section:ReviewFCN}

The pre-trained CNN is optimized for the classification purpose of low-resolution images (e.g. $224 \times 224$ pixels), whereas input images for semantic segmentation are usually in high resolution. Originally in \cite{long2015fully}, Long et al. didn't take this domain gap into consideration, so they simply adapted the architecture of pre-trained CNN in the following aspects: (1), adapt the \texttt{fc} layer to its equivalent \texttt{conv} layer (with size $1 \times 1$); (2), alter the output channel of last \texttt{fc} layer; (3), add two skip connections from low-level feature maps. Then, the shared parameters (with pre-trained CNN) are transferred to initialize FCN. Finally, FCN is fine-tuned on target segmentation datasets.
Such simple adaptation approach cannot afford to fill the domain gap, which mainly contributes to the less optimized performance of FCN.
To elaborate, the feature maps are exposed to limited contexts, so their element-wise predictions are expected to suffer from local ambiguity.
On top of pre-trained CNN, it's essential to introduce an extra module to distill context into local features.
Based on this principle, researchers have proposed different interesting solutions.

One representative branch of work is to leverage Fully Connected CRF \cite{krahenbuhl2012efficient} (FC-CRF) to contextualize the unary predictions of FCN. For example, Chen et al. \cite{chen2016deeplab} applied FC-CRF to the unary predictions of DeepLab network, and they observed obvious improved visual quality of label prediction maps on object segmentation datasets (e.g. Pascal VOC \cite{everingham2010pascal}). Subsequently, Zheng et al. \cite{zheng2015conditional} formulated FC-CRF as a Recurrent Neural Network (CRF-RNN) so that  it can be jointly learned with FCN.
Even though FC-CRF is effective towards refining the label maps, they are more like a post-processing refinement step. Thus, these works are orthogonal to our work, which can be used to further improve the performance of IFCN.

One branch of work introduces new computational layers to achieve contextual modeling, and then they integrate it with FCN to yield an end-to-end trainable segmentation network.
For example, Liu et al. \cite{liu2015semantic} adopted local convolution layers to approximate the mean field algorithm for pairwise terms in deep parsing network (DPN). Lin et al. \cite{lin2016efficient} inserted convolution layers to model the semantic compatibility between image regions.  Visin et al. \cite{Visin_2016_CVPR_Workshops} and Shuai et al. \cite{shuai2014dag}  leveraged RNNs to propagate local context in feature maps. Recently, Yu et al. \cite{yu2015multi} utilized dilated convolution kernels to expand the receptive field of feature maps.
All these computational layers are designed to encode extra context to local features, and they brought noticeable performance benefits to FCN architectures.
Similar to them, the context network in IFCN achieves context aggregation as well. Different from \cite{liu2015semantic}\cite{lin2016efficient}\cite{yu2015multi}, our context network has multiple shortcut connections, which allow it to have deeper architecture by mitigating its optimization difficulties. Thus, our context network is able to expand the receptive fields significantly larger than those networks. Moreover, shortcut connections enable IFCN to fuse rich-scale contextual predictions, whereas those network architectures don't have such property. In contrast to \cite{shuai2014dag}, IFCN is more efficient as features (in feature maps) are processed in parallel rather than sequentially as in \cite{shuai2014dag}.

Another popular architecture modification is to combine multi-scale predictions. For example, Farabet et al. \cite{farabet2013learning} and Lin et al. \cite{lin2016efficient} adopted multi-resolution input (image pyramid)  approach.
A recent pixel-level architecture - PixelNet \cite{bansal2016pixelnet} followed the hypercolumn feature approach \cite{hariharan2015hypercolumns}.
Both approaches incur either significantly higher computation time or larger memory footprints, thus they simply combine few limited contextual predictions in practice.
By contrast, IFCN adopts dense skip connections to fuse multi-scale predictions, which is a very economic as well as an effective approach to fuse very rich contextual predilections. As IFCN adopts linear fusion strategy like in \cite{long2015fully}, its performance can be further enhanced by adopting more advanced fusion method like in \cite{ghiasi2016laplacian}.


\label{Section:Review}
\section{IFCN}

\begin{figure*}
\begin{center}
  \includegraphics[width = 1.0\textwidth]{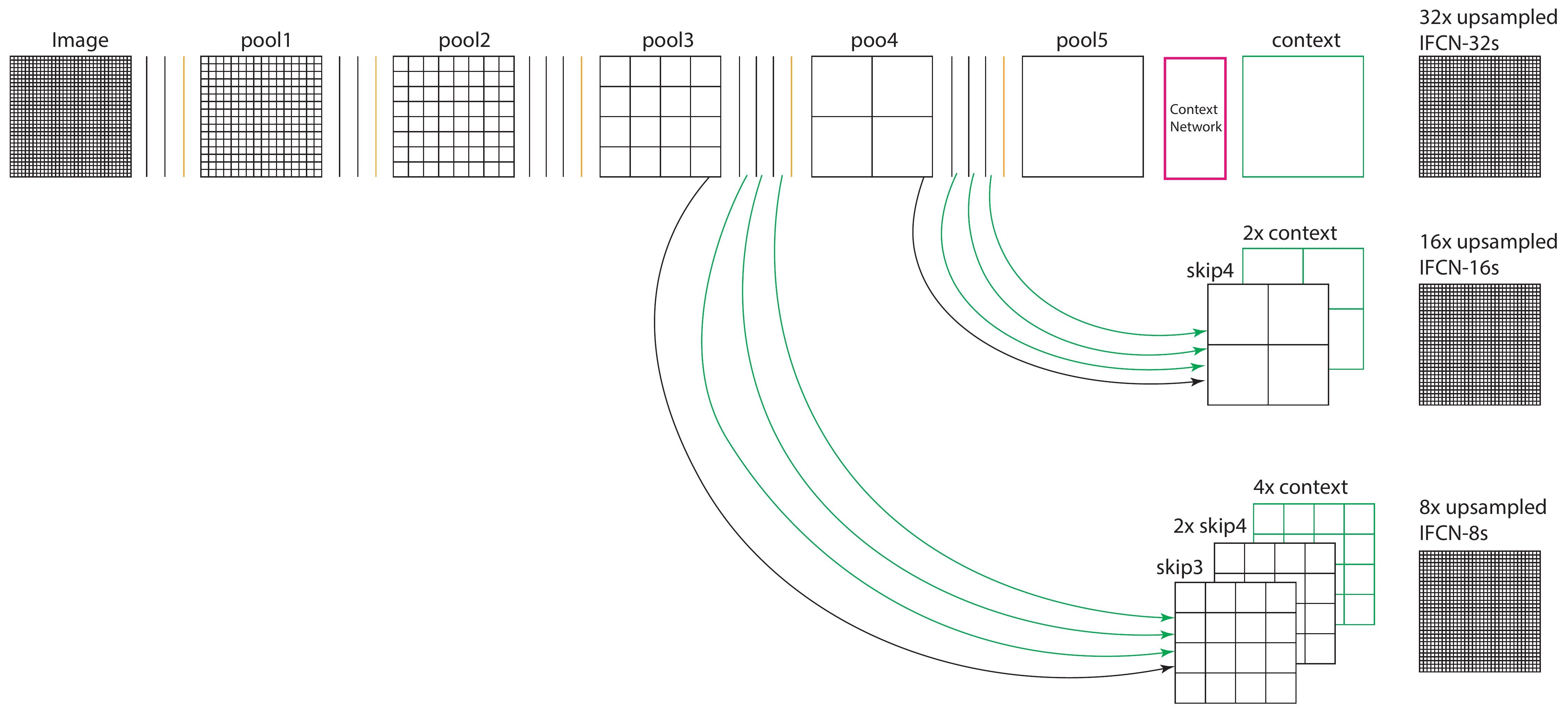}
\end{center}
\caption{Network architecture of \textbf{IFCN-xs}. Similar to the demonstration of FCN-xs, feature maps are shown as grids that reveal relative spatial coarseness. IFCN-4s and IFCN-2s can be trivially inferred from the above architecture demonstration. The black vertical line denotes the \texttt{conv} layer in the pre-trained CNN (VGG-16, c.f. Table \ref{Table:ReceptiveField-VGG-16}). In order to save space, some feature maps are not displayed. Note that the detailed architecture of context network can be retrieved in Figure \ref{Figure:ContextNetwork}. }
\label{Figure:IFCN-8s}
\end{figure*}

\subsection{Context Networks}
\label{Section:ContextNetwork}
The pre-trained CNN is optimized for the classification purpose of low-resolution images, so its network architecture is less competitive on semantic segmentation task involving high-resolution images.
Take the well-known FCN \cite{long2015fully} for demonstration, the feature map that is used for predictions (\texttt{fc7}, c.f. Table \ref{Table:ReceptiveField-VGG-16}) have a contextual view of $404 \times 404$ pixels, which is not sufficiently large for every feature to make robust prediction. Besides, previous literatures \cite{chen2016deeplab}\cite{yu2015multi} demonstrate that the contextual field of feature maps plays a critical role in the segmentation performance.
Based on such principle, we attach a context network on top of the pre-trained CNN, whose key goal is to effectively enlarge the receptive fields of feature maps.

We adopt the basic \texttt{conv} block (\texttt{Conv} + \texttt{BN} + \texttt{ReLU}) to build our context network.
The rationales behind such architecture choice are: (1), \texttt{Conv} can aggregate neighborhood context. In detail, suppose $y = k * x$, where $k$ is \texttt{Conv} kernel (with size $2m+1$), $*$ symbolizes \texttt{Conv} operation, $x$ and $y$ are 1-D input and output signals, and they have the same dimension. Mathematically, $y(i) = \sum_{j=i-m}^{j=i+m} k(i) x(i)$, thus it can be interpreted as a contextualized $x(i)$;  (2), \texttt{Conv} has been broadly and successfully devised and used in many literatures \cite{lin2016efficient}\cite{liu2015semantic}\cite{yu2015multi} to perform context aggregation; (3), \texttt{Conv} is more efficient compared to recursive / recurrent NN \cite{sharma2014recursive}\cite{shuai2014dag} to aggregate context.

Then, we are able to expand the contextual fields progressively by stacking multiple \texttt{conv} blocks. Unfortunately, the resulting segmentation network doesn't work properly according to our experiments (c.f. Section \ref{Section:SkipConnectionDiscussion}), which also aligns with the discoveries by Long et al. \cite{long2015fully}. There could be two possible reasons that lead to such undesirable behavior: (1), the network training is hard due to gradient vanishing problem when the context network is deep; (2), the final feature map generated by the context network is expected to have wide (global) range of contextual views, which makes the features less discriminative for local predictions. To address these issues, we introduce shortcut branches emanating from intermediate feature maps. The architecture of context network is shown in Figure \ref{Figure:ContextNetwork}. Specifically, those shortcut branches are responsible for locally predicting feature maps. Then these predictions are summed to generate the fused prediction map.  They can be mathematically represented as

\begin{equation}
 F = \sum_{i=1:M} \mathcal{S}(\Theta_i , f_i)
\end{equation}
where $F$ is the fused prediction map (\texttt{context} in Figure \ref{Figure:ContextNetwork}), $M$ is the quantity of \texttt{conv} blocks (i.e. depth of context network), $\mathcal{S}(\cdot, \cdot)$ represents the shortcut function, $\Theta_i$ denotes the parameters of $i$-th shortcut branch and $f_i$ is the output feature map of $i$-th \texttt{conv} block. From the equation, we can see that $f_i$ is directly connected to the error signals (exclude upsample layers), thus its preceding \texttt{conv} block receives strong supervision signals. From this perspective, the gradient vanishing problem is greatly mitigated, so the context network can be effectively trained. Moreover, the fused prediction map $F$ combines multiple decisions from $f_i$. Knowing that each $f_i$ preserves different scale context, their fused prediction map $F$ is expected to be more robust.
\subsection{Dense skip connections}
\label{Section:IFCN-8s}

FCN-8s \cite{long2015fully} adds two skip connections to FCN(-32s) so that low-level and mid-level feature maps are also utilized to make predictions. Specifically, FCN-8s fuses predictions from \texttt{pool3}, \texttt{pool4} and \texttt{fc7}. As these feature maps encode diverse scale context (c.f. Table \ref{Table:ReceptiveField-VGG-16}), it's advantageous to combine their complementary decisions to account for objects with varied scales. In practice, FCN-8s significantly outperforms FCN(-32s).

Considering that the scales of objects vary dramatically across images and scenes, we are unable to utilize a few (e.g. 3 in FCN-8s) fixed scale contexts to handle such significant scale variances. Next, we investigate the receptive fields of intermediate feature maps in pre-trained VGG-16 (FCN).  From Table \ref{Table:ReceptiveField-VGG-16}, we observe that each feature map carries distinct scale information, and they are complementary to each other. Therefore, we add skip connections from these feature maps to prediction maps, so that rich-scale contextual prediction is collectively achieved. Such architecture modification gives rise to our IFCN-xs, a similar notation to FCN-xs, where `xs' denotes that the finest resolution of prediction map has a sliding stride x. The architecture of IFCN-xs is illustrated in Figure \ref{Figure:IFCN-8s}. In comparison with FCN-xs, IFCN-xs entails more shortcut connections. These dense skip connections enable IFCN-xs to leverage very rich-scale context to make predictions, which is essential to tackle the high scale variance of objects. It's also important to note that these dense skip connections involve minor extra computation overhead comparing to FCN-xs.

In the end, IFCN-xs is trained with pixel-wise loss (similar to FCN-xs). To distribute more attention for infrequent classes, we modulate the pixel-wise loss according to its rareness magnitude as in \cite{shuai2014dag}.
This practice is economic and it is essential to significantly boost the recognition performance of rare classes.
We follow the 85\% -15\% rule to determine the rare categories, and we refer the readers to \cite{shuai2014dag} for detailed description.

\section{Architecture Evaluation on ADE20K}
\label{Section:ArchitectureEvaluation}

\noindent \textbf{ADE20K} \cite{zhou2016semantic} is a recently created large-scale dataset for ImageNet scene parsing challenge.
The dataset contains 20210 training, 2000 validation and 3352 test images.  Each pixel is annotated with one of 150 semantic categories including thing classes as well as stuff classes.
In order to modulate the rareness weighted loss \cite{shuai2014dag}, those classes whose frequency is less than $1\%$ are considered as rare. We report the results on validation images.

\vspace{0.5em}
\noindent \textbf{Experimental setups}:
The hidden dimension of \texttt{conv} blocks in context network is fixed to 512.
The new parameters engaged in the context network and skip connections are randomly initialized (Gaussian distribution with variance $10 ^{-2}$). We use convolution transpose (deconvolution) kernels \cite{long2015fully} to perform upsampling operation.
IFCN is trained with SGD with momentum (batch size 10). The learning rate is initialized to be $10^{-3}$, and it is decreased by 10 times after 15 and 20 epoches (25 epoches in total).  The momentum is fixed to 0.9. Meanwhile, higher learning rate ($3 \times$) is used for newly-initialized parameters, i.e. context network and skip connections.
In order to observe clear benefits of network architectures, we report the segmentation performance of their \textbf{unary predictions}.
In addition, IFCN is only trained with provided segmentation masks.
Images are resized to have maximum length of 512 pixels, and they are zero padded to $512 \times 512$ pixels to allow for batch processing. We randomly flip the images horizontally (on the fly) to augment the training images. The statistics (mean and variance) in batch normalization (\texttt{BN}) layer is updated after the network is converged.

\vspace{0.5em}
\noindent \textbf{Evaluation}: Three performance metrics are used to evaluate our IFCN-xs: overall pixel accuracy (\textbf{Pixel Acc.}), mean class accuracy (\textbf{Mean Acc.}) and \textbf{Mean IOU}. We refer the readers to \cite{long2015fully} for mathematical definitions.

\subsection{Architecture discussion of context network}

We evaluate several key principles for designing a good context network (i.e. densely branched convolution network in this paper).
For this purpose, we introduce the network architecture IFCN-8s-A, which has a similar structure to FCN-8s except that a context network is engaged in IFCN-8s-A. Thus, their performance gap clearly demonstrates how the introduced context network influences the segmentation behavior.

\begin{table}
\footnotesize
\centering
\begin{tabular}{lcc}
\toprule
Context network & Receptive Field (px)  & Mean IOU \\
\midrule
$3 \times 3 \ (12)$ & $980 \times 980$  & 35.67 \\
$5 \times 5 \ (6)$ & $980 \times 980$   & \textbf{35.78} \\
$7 \times 7 \ (4)$ & $980 \times 980$  & 35.11 \\
$9 \times 9 \ (3)$ & $980 \times 980$   & 34.63 \\
\midrule
FCN-8s \cite{long2015fully} & $414 \times 414$ & 29.32 \\
\bottomrule
\end{tabular}
\caption{ Performance comparisons of IFCN-8s-A (VGG-16) with different setup of context network. The format $k \times k \ (m)$ denotes that the context network is stacked with $m$ consecutive $k \times k$ convolution blocks. }
\label{Table:ContextNetworkDiscussion}
\end{table}

\vspace{0.5em}
\noindent \textbf{Architecture Shapes}.
We build multiple typical context networks, which expand the same size of receptive fields whereas they have different network depths.   A large receptive field (expanded by IFCN-8s-A) is carefully designed to make sure that nearly all elements in feature maps are accessible to full-image context. \footnote{ Note that off-centering pixels require significantly larger receptive fields than centering pixels to have the same actual contextual views. In order to make sure that all features engage full-image context, the network is expected to expand a receptive field that is significantly ($2 \times$) larger  than the size of input images.} Our preliminary experiments show that the segmentation performance of IFCN-8s-A degrades as their receptive fields go beyond this size.
We demonstrate their settings as well as the segmentation performances of their corresponding IFCN-8s-A in Table \ref{Table:ContextNetworkDiscussion}. From which, we make the following observations.

First and foremost, all IFCN-8s-A has a significantly larger receptive field than that of FCN-8s. In consequence, they outperform FCN-8s by a noticeable margin, which clearly shows the significance of context networks (receptive field expansion) inside the architecture design of segmentation networks.
In addition, deep context networks performs better than shallow alternatives when they expand equally large size of receptive fields.
This result can be interpreted from two aspects. First, the shallower network has fewer skip connections, and it generates fewer feature maps. Thus, some informative scale contextual predictions are not fused into final decisions. Meanwhile, large \texttt{conv} block (in shallow networks) refers to higher-order context aggregation, and it involves more parameters. These factors are expected to add difficulty to the training of context networks.

\vspace{0.5em}
\noindent \textbf{Batch Normalization} ( \texttt{BN} ) \cite{ioffe2015batch} is essential to accelerate training as well as to improve the segmentation performance. As depicted in Figure \ref{Figure:ContextNetwork}, we plug in a  \texttt{BN} layer  right after each convolution layer in the \texttt{conv} block. In experiments, we find that the context network with \texttt{BN} performs significantly better than that without engaging \texttt{BN}.

\subsection{Does performance improvement simply originate from engaging more parameters?}
\begin{table}[t]
\footnotesize
\centering
\begin{tabular}{lccc}
\toprule
Context network & Tied & \#Params & Mean IOU. \\
\midrule
$5 \times 5 \ (1)$ & $\times$ & 6.55M & 32.30 \\
$5 \times 5 \ (6)$ & \checkmark & 6.55M & 34.64 \\
$5 \times 5 \ (6)$ & $\times$ & 39.3M & 35.78 \\
$5 \times 5 \ (6)$ Wide & $\times$ & 144.2M & 36.08 \\
\bottomrule
\end{tabular}
\caption{Configuration of different context networks and the performance of their corresponding IFCN-8s-A (VGG-16).}
\label{Table:ParametersDiscussion}
\end{table}

First, it's crucial to introduce new parameters, which are responsible for filling the domain gap  during the fine-tuning of segmentation networks (IFCN). To illustrate this situation, let's  directly compare FCN-8s and IFCN-8s-A. In FCN-8s, the pre-trained \texttt{fc} layers (\texttt{fc6} and \texttt{fc7}) involves significantly more parameters than any context networks listed in Table \ref{Table:ContextNetworkDiscussion}. IFCN-8s-A \footnote{It's important to note that we have removed the pre-trained \texttt{fc} layers (both \texttt{fc6} and \texttt{fc7}) in IFCN (IFCN-8s-A).} (plugged in with these context networks), on the other hand, outperforms FCN-8s by a large margin.

Besides, we believe that the network architecture outweighs the magnitude of parameters in terms of boosting the performance.
To validate this claim, we conduct two controlled experiments. To be specific, we adopt the architecture of $5 \times 5 \ (m)$ to be the basic component of context networks, where $m$ denotes the quantity of \texttt{conv} blocks (i.e. network depth). We further plug them into IFCN-8s-A, and compare their segmentation performance in Table \ref{Table:ParametersDiscussion}. We interpret the results from two perspectives.

\vspace{0.5em}
\noindent \textbf{Tied Parameters}. First, we discuss the context network with tied parameters. Under this scenario, the amount of parameters engaged by context networks is significantly reduced. 
In our experiment, the tied $5 \times 5 \ (6)$ network has the same amount of parameters with $5 \times 5 \ (1)$. However, the former network largely outperforms the $5 \times 5 \ (1)$ alternative, as the tied $5 \times 5 (6)$ network expand significantly larger receptive fields than $5 \times 5(1)$. This result convinces us to believe that the huge performance boost mainly results from the architecture of context networks.

\vspace{0.5em}
\noindent \textbf{Wide context network}. Then, we train a wide network whose architecture configuration is $5 \times 5 \ (6)$. Different from previous settings, the hidden dimension of wide network is increased from $512$ to $1024$. Under this scenario, the wide network involves magnitude larger parameters.  However, it fails to generate significantly better result.
Therefore, it's fair to conclude that the increase of network parameters is not the key contributor towards performance boost.

\subsection{IFCN-xs: Is dense skip connection necessary ?}
\label{Section:SkipConnectionDiscussion}

\begin{table}[t]
\footnotesize
\centering
\begin{tabular}{lccc}
\toprule
Network & Pixel Acc. & Mean Acc. & Mean IOU \\
\midrule
IFCN-32s & 73.67 & 48.40  & 33.17 \\
IFCN-16s & 75.44 & 51.53  & 36.26 \\
IFCN-8s & \textbf{76.03} & \textbf{52.00} & \textbf{36.98} \\
IFCN-4s & 75.78 & 51.31 & 36.02 \\
\midrule
IFCN-8s-A & 75.37 & 50.60 & 35.78 \\
IFCN-8s-B & 73.06 & 43.91 & 31.18 \\
\bottomrule
\end{tabular}
\caption{Performance comparisons of different IFCN (VGG-16) architectures. Please refer to the text for description.}
\label{Table:DenseConnectionDiscussion}
\end{table}

To answer this question, we investigate three parts of shortcut connections in IFCN-xs.

\vspace{0.5em}
\noindent \textbf{IFCN-8s-A vs IFCN-8s-B}. We first build IFCN-8s-B by removing the shortcut branches in IFCN-8s-A. Then we directly compare their segmentation performance. As shown in Table \ref{Table:DenseConnectionDiscussion}, the performance of IFCN-8s-B significantly lags behind that of IFCN-8s-A. Such a performance gap clearly demonstrates that the shortcut branches indeed improves the training of context networks. Besides, it also indicates that fusion of multiple scale contextual predictions is beneficial.

\vspace{0.5em}
\noindent \textbf{IFCN-8s vs IFCN-8s-A}. Next, we compare these two network architectures. The architecture of FCN-8s is presented in Figure \ref{Figure:IFCN-8s}. IFCN-8s-A follows the architecture design of FCN-8s \cite{long2015fully}. In Table \ref{Table:DenseConnectionDiscussion}, we observe a noticeable performance improvement when we switch IFCN-8s-A to IFCN-8s. This result shows that the dense shortcut connections enable IFCN to fuse rich-scale contextual predictions, which is crucial towards achieving high performance.

\vspace{0.5em}
\noindent \textbf{IFCN-xs}. Finally, we empirically discuss which skip connection should be included in the IFCN-xs architecture. Considering that it's extremely expensive as well as unnecessary to evaluate every skip connection, we simply evaluate the skip connections in one block. In more details, the feature maps are considered to be in the same block as long as they have the same spatial resolution. Take VGG-16  in Table \ref{Table:ReceptiveField-VGG-16} as an example, feature maps \texttt{pool4}, \texttt{conv5\_1}, \texttt{conv5\_2}, \texttt{conv5\_3} have the same resolution, so skip connections emanating from them are deemed as in the same block. By adding these skip connections progressively, we are able to produce a number of IFCN-xs whose architecture is illustrated in Figure \ref{Figure:IFCN-8s}. Among all the architectures, IFCN-8s performs the best. Though IFCN-4s entails more skip connections than IFCN-8s, it however achieves inferior segmentation results. We believe that the feature maps attached by the added skip connections (in IFCN-4s, namely \texttt{pool2}, \texttt{conv3\_1}, \texttt{conv3\_2}, \texttt{con3\_3}) incorporate too limited context (less than $40 \times 40$ pixels, c.f. Table \ref{Table:ReceptiveField-VGG-16}), thus the predictions based on them is expected to be noisy. It's empirically not advisable to include these skip connections in the IFCN architecture.

\subsection{Can the proposed architecture work for deeper pre-trained network?}
\begin{table}[t]
\footnotesize
\centering
\begin{tabular}{lccc}
\toprule
\multicolumn{1}{c}{Methods} & \multicolumn{3}{c}{ResNet-50}  \\
\midrule
FCN-8s & 74.42 & 47.42 & 33.89 \\
IFCN-8s & \textbf{76.97} & \textbf{53.17} & \textbf{38.06} \\
\bottomrule
\end{tabular}


\begin{tabular}{lccc}
\toprule
\multicolumn{1}{c}{Methods} & \multicolumn{3}{c}{ResNet-101}  \\
\midrule
FCN-8s & 75.96 & 50.11 & 35.76 \\
IFCN-8s & \textbf{77.74} & \textbf{55.39} & \textbf{39.73} \\
\bottomrule
\end{tabular}
\caption{Performance (Pixel Acc. , Mean Acc. Mean IOU) of different network architectures (with different pre-trained CNNs) on ADE20K dataset \cite{zhou2016semantic}.  }
\label{Table:DeepNetworkDiscussion}
\end{table}

Due to the recent success of residue learning \cite{he2016deep}, this question arises naturally as we are easily accessible to very deep pre-trained CNN nowadays.
For example, the pre-trained deep networks (e.g. ResNet-101) have already expanded very large receptive fields.
In this scenario, it's curious and important to know whether the proposed architecture design (IFCN) is still able to bring performance benefits, if not as significant as that in VGG-16.

We adapt the pre-trained ResNets (ResNet-50 \cite{he2016deep} and ResNet-101 \cite{he2016deep}) to two different architectures: FCN-8s and IFCN-8s. As shown in Table \ref{Table:DeepNetworkDiscussion}: IFCN-8s consistently and significantly outperforms FCN-8s, which again validates the solid contributions of this paper. This result is inspiring and interesting considering that FCN-8s adapted from deep ResNet is already strong.
Knowing that the pre-trained ResNets are trained from low-resolution images, the introduced context network  is essential and effective to adapt the feature maps so that they are optimized for the segmentation purpose of high-resolution images. Meanwhile, the fusion of rich-scale contextual predictions achieved by dense skip connections also contribute significantly to the performance boost.

\section{Results on segmentation benchmarks}
\label{Section:ResultsSegmentation}

We evaluate our IFCN-8s on standard semantic segmentation datasets, and the architecture of its context network  is $5 \times 5 (6)$.  The same experimental setups (as in controlled experiments, c.f. Section \ref{Section:ArchitectureEvaluation}) are used to train IFCN-8s over different datasets. It's worth mentioning that wider context network has higher capacity, and it can further generate slightly better performance (c.f. Table \ref{Table:ParametersDiscussion}).

\begin{table}[t]
\footnotesize
\centering
\begin{tabular}{lccc}
\toprule
Methods & Pixel Acc. & Mean Acc. & Mean IoU \\
\midrule
IFCN-8s (VGG-16)  & \textbf{76.03} & \textbf{52.00} & \textbf{36.98} \\
IFCN-8s (ResNet-101) & \textbf{77.74} & \textbf{55.39} & \textbf{39.73} \\
\midrule
FCN-8s \cite{long2015fully} & 71.32 & 40.32 & 29.32 \\
SegNet \cite{badrinarayanan2015segnet} & 71.00 & 31.14 & 21.64 \\
DilatedNet \cite{yu2015multi} & 73.55 & 44.59 & 32.31 \\
\bottomrule
\end{tabular}
\caption{\textbf{ADE20K} (150 classes) validation accuracies (\%). All methods except IFCN-8s (ResNet-101) use VGG-16.}
\label{Table:resultsImageNet}
\end{table}

\vspace{0.5em}
\noindent \textbf{ADE20K results} are listed in Table \ref{Table:resultsImageNet}.

\begin{table}[t]
\footnotesize
\centering
\begin{tabular}{lccc}
\toprule
Methods & Pixel Acc. & Mean Acc. & Mean IoU \\
\midrule
\emph{\textbf{VGG-16}} & & & \\
IFCN-8s& \textbf{74.5} & \textbf{57.7} & \textbf{45.0} \\
\midrule
FCN-8s \cite{long2015fully}  & 67.0 & 50.7 & 37.8 \\
DeepLab \cite{chen2016deeplab} & n/a & n/a & 37.6 \\
DeepLab + CRF \cite{chen2016deeplab} & n/a & n/a & 39.6 \\
CRF-RNN \cite{zheng2015conditional} & n/a & n/a & 39.3 \\
ParseNet \cite{liu2015parsenet} & n/a & n/a & 40.4 \\
ConvPP-8 \cite{xie2015convolutional} & n/a & n/a & 41.0 \\
PixelNet \cite{bansal2016pixelnet} & n/a & 51.5 & 41.4 \\
UoA-Context + CRF \cite{lin2016efficient} & 71.5 & 53.9 & 43.3 \\
\bottomrule
\toprule
\emph{\textbf{ResNet-101}} & & & \\
IFCN-8s & \textbf{77.5} & \textbf{62.6} & \textbf{49.9} \\
DeepLab \cite{chen2016deeplab} & n/a & n/a & 41.4 \\
FCRN \cite{wu2016bridging} & 72.9 & 54.8 & 44.5 \\
\bottomrule
\end{tabular}
\caption{\textbf{Pascal Context} (59 classes) validation accuracies (\%).}
\label{Table:resultsContext}
\end{table}

\vspace{0.5em}
\noindent \textbf{Pascal Context results} are shown in Table \ref{Table:resultsContext}.
Pascal Context \cite{mottaghi_cvpr14} has 10103 images, out of which 4998 images are used for training. The images are from Pascal VOC 2010 datasets, and they are re-labeled as pixel-wise segmentation maps which include 540 semantic classes (including the original 20 classes). Similar to Mottaghi et al. \cite{mottaghi_cvpr14}, we only consider the most frequent 59 classes in the dataset for evaluation.  Based on the rareness identification rule in \cite{shuai2014dag}, those classes whose frequencies are lower than $1\%$  are identified as rare.

\begin{figure}[t]
\centering
  \includegraphics[width=0.5\textwidth]{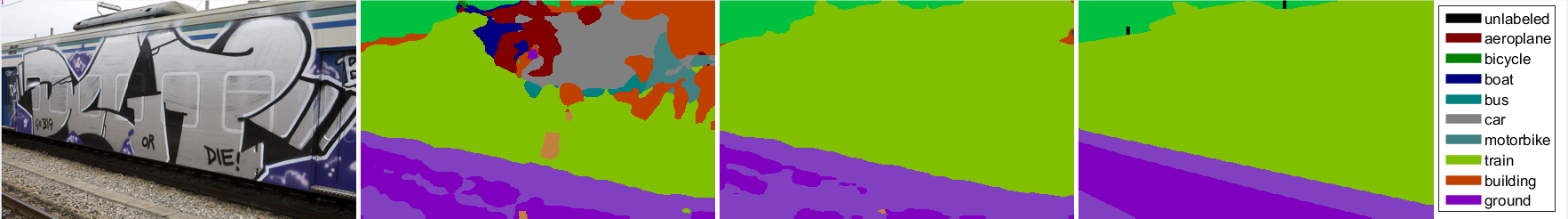}
  \includegraphics[width=0.5\textwidth]{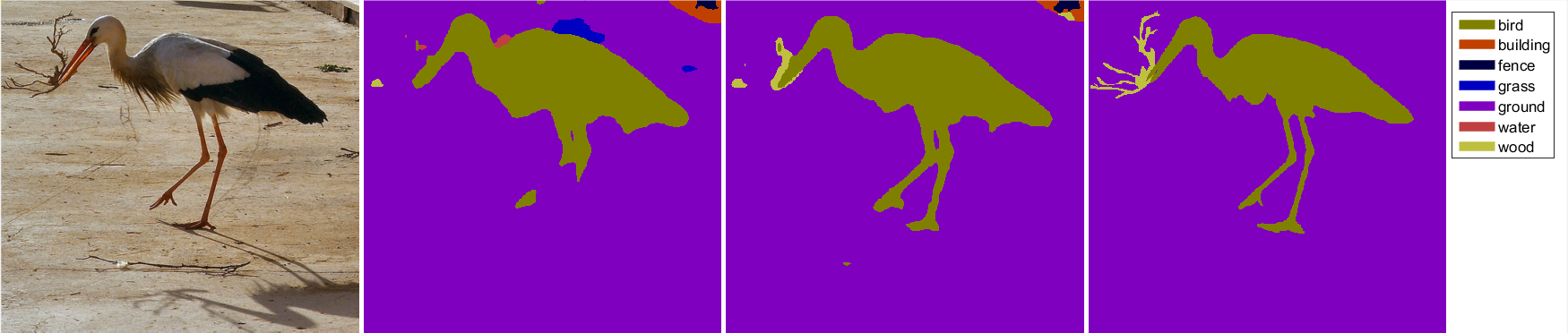}
  \includegraphics[width=0.5\textwidth]{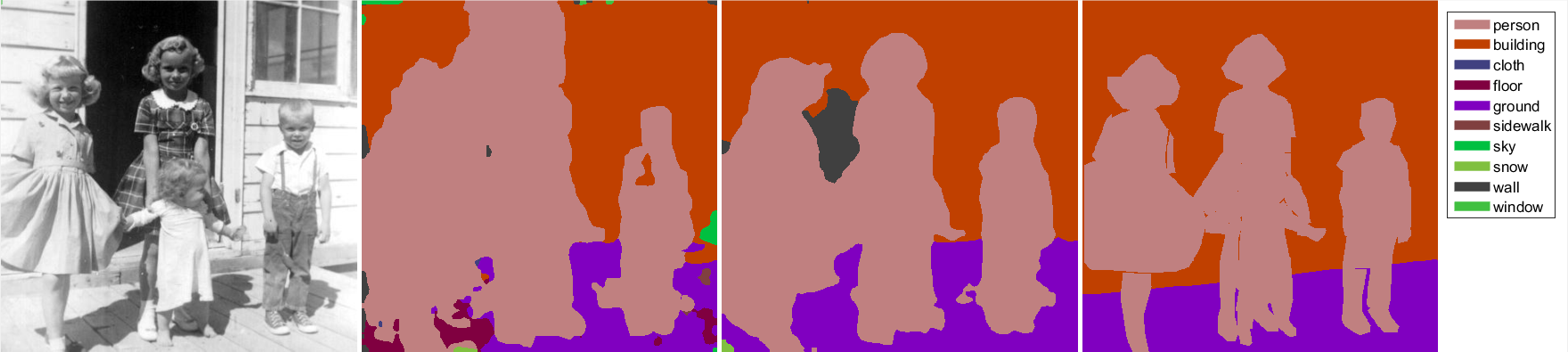}
  \includegraphics[width=0.5\textwidth]{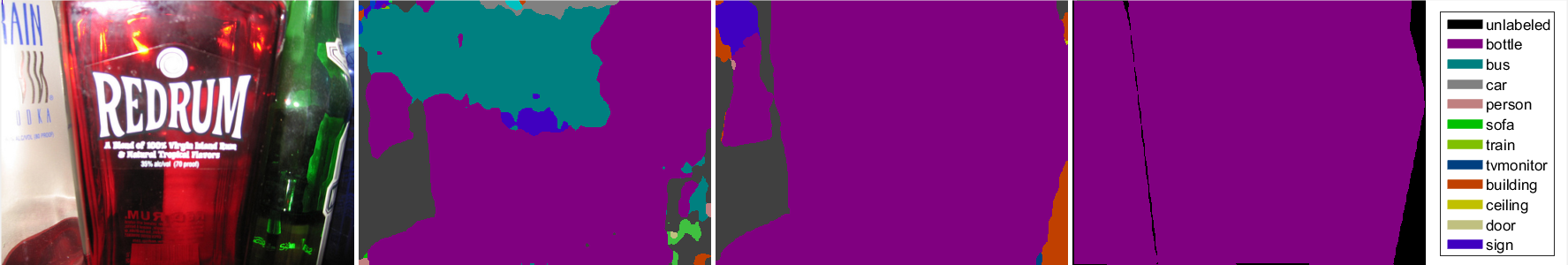}
  \includegraphics[width=0.5\textwidth]{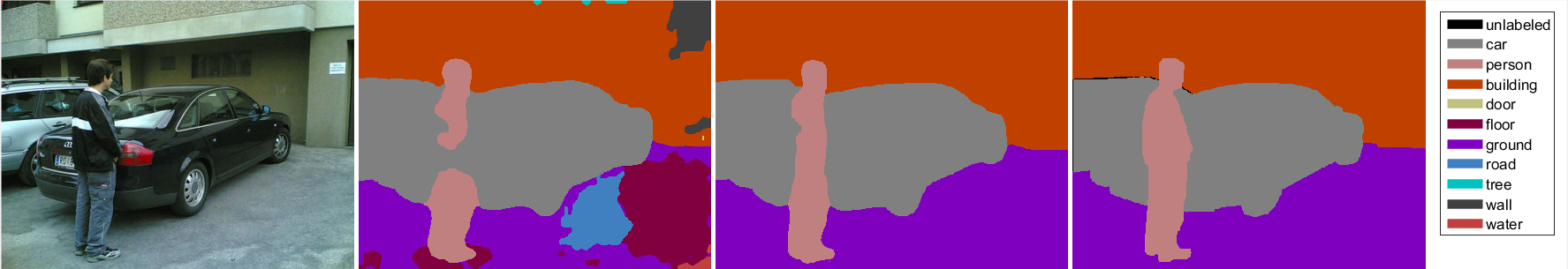}
\caption{Qualitative segmentation result comparisons. In each row, we show \textbf{input images}, unary prediction maps from \textbf{FCN-8s, IFCN-8s (VGG-16)} and \textbf{ground truth}. The figure is best viewed in electronic version of this paper with 400\% zoom-in. Please refer to texts in Section \ref{Section:ResultSummary} for detailed elaboration.}
\label{Figure:QualitativeResults}
\end{figure}

\begin{table}[t]
\footnotesize
\centering
\begin{tabular}{lccc}
\toprule
Methods  & Pixel Acc. & Mean Acc. & Mean IOU \\
\midrule
IFCN-8s (VGG-16)  & \textbf{76.90} & \textbf{53.46} & \textbf{40.74} \\
IFCN-8s (ResNet-101)  &\textbf{77.70} & \textbf{55.46} & \textbf{42.69} \\
\midrule
FCN \cite{long2015fully}  & 68.18 & 38.41 & 27.39 \\
DeconvNet \cite{noh2015learning}  & 66.13 & 33.28 & 22.57 \\
SegNet \cite{badrinarayanan2015segnet}  & 72.63 & 44.76 & 31.84 \\
DeepLab \cite{chen2016deeplab} & 71.90 & 42.21 & 32.08 \\
\bottomrule
\end{tabular}
\caption{\textbf{SUN-RGBD} (37 classes) testing accuracies (\%).  We only use RGB modality in our experiments. All methods except IFCN-8s(ResNet-101) use VGG-16, and all other reported results are copied from \cite{badrinarayanan2015segnet}. }
\label{Table:resultsSUNRGBD}
\end{table}

\vspace{0.5em}
\noindent \textbf{SUN-RGBD Results} are reported in Table \ref{Table:resultsSUNRGBD}.
SUN-RGBD \cite{song2015sun} contains images from NYU depth V2 \cite{silberman2012indoor}, Berkeley B3DO \cite{janoch2013category}, SUN3D \cite{xiao2013sun3d} as well as the newly captured images. It has 10335 images in total, out of which 5285 images are used for training. The rareness frequency threshold is fixed to $2.5\%$ based on the 85\%-15\% rule \cite{shuai2014dag}. We follow previous literatures \cite{badrinarayanan2015segnet} to consider 37 classes for evaluation. Note that we only use RGB modality as input.

\begin{table}[t]
\footnotesize
\centering
\begin{tabular}{lc}
\toprule
Methods & Mean IOU. \\
\midrule
\emph{\textbf{VGG-16}} & \\
IFCN-8s & 74.6  \\
IFCN-8s + CRF & 75.3 \\
\midrule
FCN-8s \cite{long2015fully} & 62.2  \\
Zoom-out \cite{mostajabi2015feedforward} & 69.6\\
PixelNet \cite{bansal2016pixelnet} & 69.7 \\
DeepLab-v1 + CRF \cite{chen2015semantic} & 71.6 \\
CRF-RNN \cite{zheng2015conditional} & 72.0 \\
DeconvNet \cite{noh2015learning} & 72.5 \\
$^\dagger$DilatedNet \cite{yu2015multi} & 73.5  \\
DPN \cite{liu2015semantic} & 74.1  \\
UoA-Context + CRF\cite{lin2016efficient} & 75.3 \\
LRR \cite{ghiasi2016laplacian} & 74.7 \\
LRR + CRF \cite{ghiasi2016laplacian} & \textbf{75.9} \\
\bottomrule
\toprule
\emph{\textbf{ResNet-101} }&  \\
IFCN-8s & 79.3\\
IFCN-8s + CRF & \textbf{80.0} \\
\midrule
$^\dagger$ FCRN \cite{wu2016bridging} & 79.1 \\
$^\dagger$ LRR + CRF \cite{ghiasi2016laplacian} & 79.3 \\
$^{\dagger}$DeepLab-v2  + CRF \cite{chen2016deeplab} & 79.7 \\
\bottomrule
\end{tabular}
\caption{\textbf{Pascal VOC 2012} testing mean IOU (\%). The preceding  symbol {$^{\dagger}$} indicates that the network is pre-trained with COCO \cite{lin2014microsoft} dataset. \textcolor{blue}{\textbf{Note that COCO pre-training is not used to improve the predictions of IFCN-8s.}}}
\label{Table:resultsVOC}
\end{table}

\vspace{0.5em}
\noindent \textbf{Pascal VOC 2012 results} are presented in Table \ref{Table:resultsVOC}.
Pascal VOC 2012 originally contains 2913 train and validation images,  1456 testing images. Each pixel belongs to one of the pre-defined object or background categories.  We follow \cite{long2015fully} to augment the training set from \cite{hariharan2014simultaneous}. Thus, we end up with having 12031 training images. The rareness frequency threshold of is set to $1.5\%$. We submit our prediction maps to the evaluation servers to get the reported testing results. Detailed class-wise IOU (Jaccard) score can be retrieved in supplementary materials \footnote{They can be downloaded from the author's homepage.}.

\subsection{Result Summary}
\label{Section:ResultSummary}

\noindent \textbf{IFCN vs FCN}.
IFCN-8s demonstrates significantly ($> 7.5\% $ mIOU) better quantitative results than FCN-8s on all testing segmentation benchmarks. It's important to note that these datasets cover wide range of scenarios that include object segmentation \cite{everingham2010pascal}, outdoor scene parsing \cite{zhou2016semantic}\cite{mottaghi_cvpr14} as well as indoor scene labeling \cite{zhou2016semantic}\cite{song2015sun}. Thus, it's fair to conclude that IFCN is a versatilely better segmentation network architecture than FCN.

Next, we qualitatively compare their segmentation maps in Figure \ref{Figure:QualitativeResults}. As shown, FCN-8s struggles to make robust predictions for zoom-in objects (e.g. \nth{1} and \nth{4} example ), as it has limited receptive fields. IFCN-8s can reliably correct those errors. In addition, IFCN-8s is able to produce very detailed segmentation maps that have sharper boundaries than FCN-8s (e.g. \nth{2} and \nth{3} examples). We speculate that the dense skip connections originating from early layers of pre-trained CNNs play a significant role to achieve such appealing property of IFCN.

We further compare the inference time of IFCN-8s with FCN-8s in Table \ref{Table:InferenceTime}. As shown, IFCN-8s (VGG-16) takes reasonably longer time than FCN-8s to generate the prediction maps.
We believe that such time consumption is competitive when it is positioned with other recent segmentation network architectures, e.g. DeepLab \cite{chen2016deeplab}, DilatedNet \cite{yu2015multi} etc.
The reported time statistics are based on the average feed-forward time for images in Pascal Context dataset \cite{mottaghi_cvpr14} under a single Titan X (exclude the time consumption for image pre-processing).

\vspace{0.5em}
\noindent \textbf{IFCN vs State-of-the-arts}. IFCN has also produced significantly better performance over other recently developed segmentation network architectures including DilatedNet \cite{yu2015multi}, ParseNet \cite{liu2015parsenet}, DeepLab \cite{chen2016deeplab}, PixelNet \cite{bansal2016pixelnet}, SegNet \cite{badrinarayanan2015segnet}, etc.
In the setting of VGG-16, the unary predictions of IFCN-8s already outperforms state-of-the-arts by a large margin on ADE20K, Pascal Context and SUN RGB-D datasets.
On Pascal VOC 2012, IFCN-8s (+ CRF \cite{krahenbuhl2012efficient}) also ranks highly competitive among all of top methods.
In the advanced setting where pre-trained CNN is initialized with ResNet-101 \cite{he2016deep}, IFCN further significantly advances state-of-the arts on all datasets. It's important to note that IFCN has not been pre-trained with auxiliary segmentation data (e.g. Microsoft COCO dataset \cite{lin2014microsoft}) in all cases.

\begin{table}[t]
\footnotesize
\centering
\begin{tabular}{lcc}
\toprule
Network & Inference Time (ms) & Mean IOU (\%)\\
\midrule
FCN-8s \cite{long2015fully} & 25.0 & 37.8 \\
IFCN-8s (VGG-16) & 47.9 & 45.0 \\
IFCN-8s (ResNet-101) & 167.3 & 49.9 \\
\bottomrule
\end{tabular}
\caption{The average inference time (for images in the Pascal Context dataset) of different network architectures. }
\label{Table:InferenceTime}
\end{table}

\section{Conclusion}
In this paper, we discuss the possible limitations of \emph{de facto} segmentation network - FCN. To address these architecture flaws, we present our Improved FCN (IFCN), which is a strong segmentation network architecture that marries effective context aggregation as well as rich-scale contextual predictions in an elegant framework. In detail, a context network is attached on top of pre-trained CNN to expand the receptive fields of feature maps. In addition, dense shortcut connections are added to enable IFCN to fuse very rich contextual predictions. Empirically, IFCN consistently and significantly outperforms FCN  on standard semantic segmentation datasets including ADE20K, Pascal Context, Pascal VOC2012 and SUN-RGBD. Moreover, IFCN achieves new state-of-the-arts on all these datasets.


{\footnotesize
\bibliographystyle{ieee}
\bibliography{cvpr-2017}
}

\end{document}